\documentclass[letterpaper, 10 pt, conference]{ieeeconf}  % Comment this line out if you need a4paper

\IEEEoverridecommandlockouts                              % This command is only needed if 
\usepackage{amsmath}
\usepackage{graphicx}
\usepackage[caption=false,font=footnotesize]{subfig}
\title{\LARGE \bf
Goal-Directed Behavior under Variational Predictive Coding: \\Dynamic Organization of Visual Attention and Working Memory
}
\author{Minju Jung$^{1,2}$, Takazumi Matsumoto$^{2}$ and Jun Tani$^{2,*}$% <-this % stops a space
\thanks{$^{*}$Jun Tani is a corresponding author ({\tt\small jun.tani@oist.jp}).}%
\thanks{$^{1}$Korea Advanced Institute of Science and Technology, Daejeon, Korea}%
\thanks{$^{2}$Okinawa Institute of Science and Technology, Okinawa, Japan}%
}

\begin{document}

\maketitle
\thispagestyle{empty}
\pagestyle{empty}

%%%%%%%%%%%%%%%%%%%%%%%%%%%%%%%%%%%%%%%%%%%%%%%%%%%%%%%%%%%%%%%%%%%%%%%%%%%%%%%%
\begin{abstract}
Mental simulation is a critical cognitive function for goal-directed behavior because it is essential for assessing actions and their consequences. 
When a self-generated or externally specified goal is given, a sequence of actions that is most likely to attain that goal is selected among other candidates via mental simulation.
Therefore, better mental simulation leads to better goal-directed action planning. 
However, developing a mental simulation model is challenging because it requires knowledge of self and the environment. 
The current paper studies how adequate goal-directed action plans of robots can be mentally generated by dynamically organizing top-down visual attention and visual working memory. 
% Particularly, we examined top-down visual attention and visual working memory better mental simulation. 
For this purpose, we propose a neural network model based on variational Bayes predictive coding, where goal-directed action planning is formulated by Bayesian inference of latent intentional space. 
% Specifically, select an appropriate sequence of actions that produces a desired outcom through mental simulation. 
% The optimal goal-directed visuomotor planning is formulated by Bayesian inference of latent intentional states in which time-dependent variables of visual attention filter as well as pixel-wise mask of visual working memory at each step can be estimated. 
Our experimental results showed that cognitively meaningful competencies, such as autonomous top-down attention to the robot end effector (its hand) as well as dynamic organization of occlusion-free visual working memory, emerged. 
% Also, visual attention completely different depend on a given goal.
Furthermore, our analysis of comparative experiments indicated that introduction of visual working memory and the inference mechanism using variational Bayes predictive coding significantly improve the performance in planning adequate goal-directed actions.

% The current paper studies how adequate goal-directed action plans of robots can be generated by dynamically organizing top-down visual attention and visual working memory.
% It has been 
% For this purpose, we propose a neural network model based on variational predictive coding. 
% The proposed model encodes high-dimensional visuomotor sequences to low-dimensional latent intentional space.
% However, unlike previous predictive coding based models, 
% Therefore, goal-directed action planning is formulated by Bayesian inference of latent intentional space.
% Specifically, select an appropriate sequence of actions that produces a desired outcom through mental simulation. 
% % The optimal goal-directed visuomotor planning is formulated by Bayesian inference of latent intentional states in which time-dependent variables of visual attention filter as well as pixel-wise mask of visual working memory at each step can be estimated. 
% The experiment results showed that cognitively meaningful competency emerges such as autonomous top-down attention to the robot end effector as well as dynamic organization of occlusion-free visual working memory. Our analysis on comparative experiments indicated that introduction of visual working memory and the inference mechanism using variational Bayes significantly improve the performance in generating adequate goal-directed actions.
\end{abstract}

%%%%%%%%%%%%%%%%%%%%%%%%%%%%%%%%%%%%%%%%%%%%%%%%%%%%%%%%%%%%%%%%%%%%%%%%%%%%%%%%
\section{Introduction}
Goal-directed behavior is the ability to generate optimal action plans to achieve goals and to execute the action plans generated. %planning and executing appropriate actions to achieve goals.
Understating how humans recognize goals of others by observation and translate the goals into goal-directed actions is a crucial step toward general intelligence. %humans translate goals into goal-directed behavior and recognizing the goals of others through their behavior is a key step toward general intelligence.
Goal-directed behavior is becoming increasingly important in the robotics literature because future robots, such as collaborative and home assistant robots, need to execute a variety of goal-directed behaviors, rather than a single fixed one \cite{Finn:2017,Choi:2018,Nair:2018}.
% For example, collaborative robots .
However, even a simple goal-directed behavior for robots, such as grasping a cup, is not a trivial problem, because the location and shape of a cup need to be extracted visually, and hand-eye coordination must be programmed or learned before action execution \cite{Levine:2018}.
% How can we build intelligent robots that are able to ?

\begin{figure}[!t]
\centering
\includegraphics[width=0.45\textwidth]{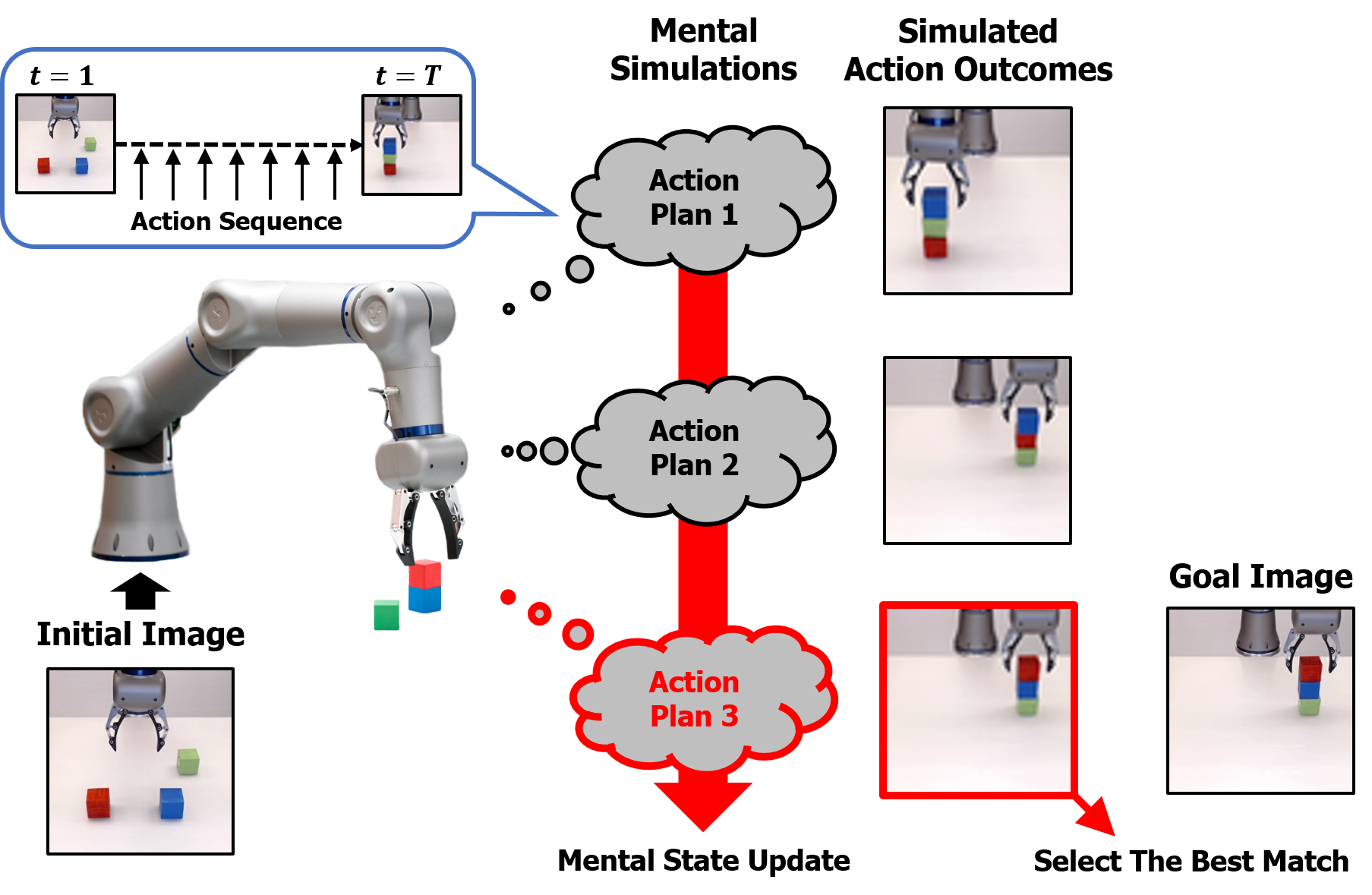}
%\hfil
\vspace{-2mm}
\caption{Goal-directed action planning via iterative mental simulation. A robot mentally simulates different courses of action by updating a robot's mental state and then selects a plan based on the simulated action outcome, which is most similar to the desired goal. However, mental simulation becomes more challenging as the length of a plan $T$ increased.}
%(c) Comparison between the baseline and AD when the update gate bias is initialized to -2.
\label{fig:teaser}
\vspace{-6mm}
\end{figure}

The capacity to anticipate action outcomes based on knowledge of the causal structure of the environment, which is known as mental simulation, is crucial for goal-directed behavior in the brain \cite{Grezes:2001,Pezzulo:2014a}. %Pezzulo:2014b,
% By doing so, 
In a similar vein, one promising approach in the robotics literature to address goal-directed behavior is based on predictive models (Fig. \ref{fig:teaser}).
% The predictive coding framework is (\cite{}). There are number of evidences 
Finn and Levine \cite{Finn:2017} showed that predictive models can be used for robot planning by selecting an action sequence that maximizes the probability of the desired outcome. However, the following limitations exist: (1) the model was trained to predict vision, but not action, (2) actions were estimated by a stochastic optimization algorithm, which is not part of the predictive model, and (3) a plan was estimated within a short period of time.
Choi et al. \cite{Choi:2018} proposed a neural network model based on the predictive coding framework \cite{Rao:1999,Friston:2010}. 
The model was trained to encode high-dimensional visuomotor sequences into corresponding initial states in low-dimensional latent intentional space. 
By doing so, planning can be defined as finding an initial state that mentally generates a visuomotor sequence to achieve a given goal. 
Therefore, mental simulation is a key capacity to plan effectively. 
% By doing so, the model can encode the latent visuomotor features of each training sample into the corresponding initial state, which is considered an intention.
% The network was trained to predict visuomotor sequences jointly by updating the parameters of the network and the initial states corresponding to each training sample. 
% By doing so, the model can encode the latent visuomotor features of each training sample into the corresponding initial state, which is considered an intention.
However, predicting high-dimensional visuomotor sequences becomes more challenging as sequences become longer, even though it is essential for complex goal-directed action planning. 
Furthermore, Choi et al. \cite{Choi:2018} reported that deterministic predictive coding requires a large number of training samples for good generalization.
%and (2) goal representations are required \cite{}.
How can we address long-term visuomotor prediction and generalization problems in the predictive coding framework for better goal-directed action planning?

Humans have two kinds of vision: central vision and peripheral vision.
Central vision allows us to focus on important visual stimuli from a narrow receptive field at high resolution, while peripheral vision has a wide receptive field at low resolution.
By coordinating the two complementary vision types, humans are able to deal with a high-dimensional raw visual stream efficiently.
Especially, goal-directed behavior focuses visual attention on goal-relevant stimuli and ignores irrelevant stimuli in a top-down manner.
% visual attention of central vision shifts continuously across the visual field in order to extract goal-relevant visual information in a top-down manner. 
Visual information perceived by the retina is first stored in iconic memory, but it rapidly disappears \cite{Coltheart:1980}.
Therefore, some visual information must be transferred to visual working memory for information integration across visual attention shifts and future usage.
% Therefore, the visual information that is necessary for future use should be transferred to visual working memory. % (i.e. visuospatial sket/chpad), which is a key component of working memory. % for maintenance and manipulation. 
Working memory is a cognitive brain system that temporarily stores and manipulates an active representation of information to provide an interface between perception, long-term memory, and action \cite{Baddeley:2003}. 
Visual working memory is the specific type of working memory for visual information and involves mental simulation or manipulation \cite{Sims:1997,Roelfsema:2016}. 
% Furthermore, 
% Visual working memory stores visual information and can mentally manipulate internally stored visual information.
% % mental simulation that supports to understanding, reasoning, and predicting the behavior of physical systems.
%  visual attention and working memory constitute a single system. 

In this paper, we employ top-down visual attention and visual working memory to address the difficulty of long-term visuomotor prediction in the predictive coding framework.
The proposed model has two visuomotor stream sub-networks: dorsal and ventral. The segregation of the dorsal and ventral visuomotor streams is based on the source of the visual information. 
Dividing high-dimensional visual processing into low-dimensional peripheral and central visual processing reduces the degree of difficulty of long-term visual prediction and the computational burden for training and planning. 
For visual prediction, we propose a visual prediction module that merges peripheral and central visual prediction into a single visual prediction with the help of an external visuospatial memory, called background memory, working as visual working memory.
Background memory preserves long-term visuospatial information, which is crucial for long-term action planning.

% Furthermore, variational Bayes is employed to enhance the generalization capability. 
Since the proposed model is under the predictive coding framework, the proposed model is trained to generate multiple visuomotor sequences depending on the corresponding initial state. However, unlike a deterministic predictive coding framework, we incorporate a variational Bayes approach \cite{Kingma:2013} into predictive coding framework, called  variational predictive coding, to enhance generalization capability with a small number of training samples.
In the block stacking experiment, the results showed that the proposed model can plan appropriate visuomotor sequences that achieve goals by updating initial states in the direction of prediction error minimization. 
Without any supervision, smooth pursuit tracking of the end effector by top-down visual attention as well as maintenance and manipulation in background memory is self-organized and varied depending on goals. %smooth pursuit-like eye movements to track its hand,
% In the qualitative analysis, we found that visual attention feeding to central vision follows the end effector of the robot and background memory maintains and updates block configuration. 
Also, the proposed model outperformed, in terms of visual and motor prediction, baseline models: the model without background memory and the model based on deterministic predictive coding. It indicates that the background memory and variational predictive coding are essential for complex goal-directed action planning.

\section{Method}
\begin{figure}[!t]
\centering
    \subfloat[]{\includegraphics[width=0.4\textwidth]{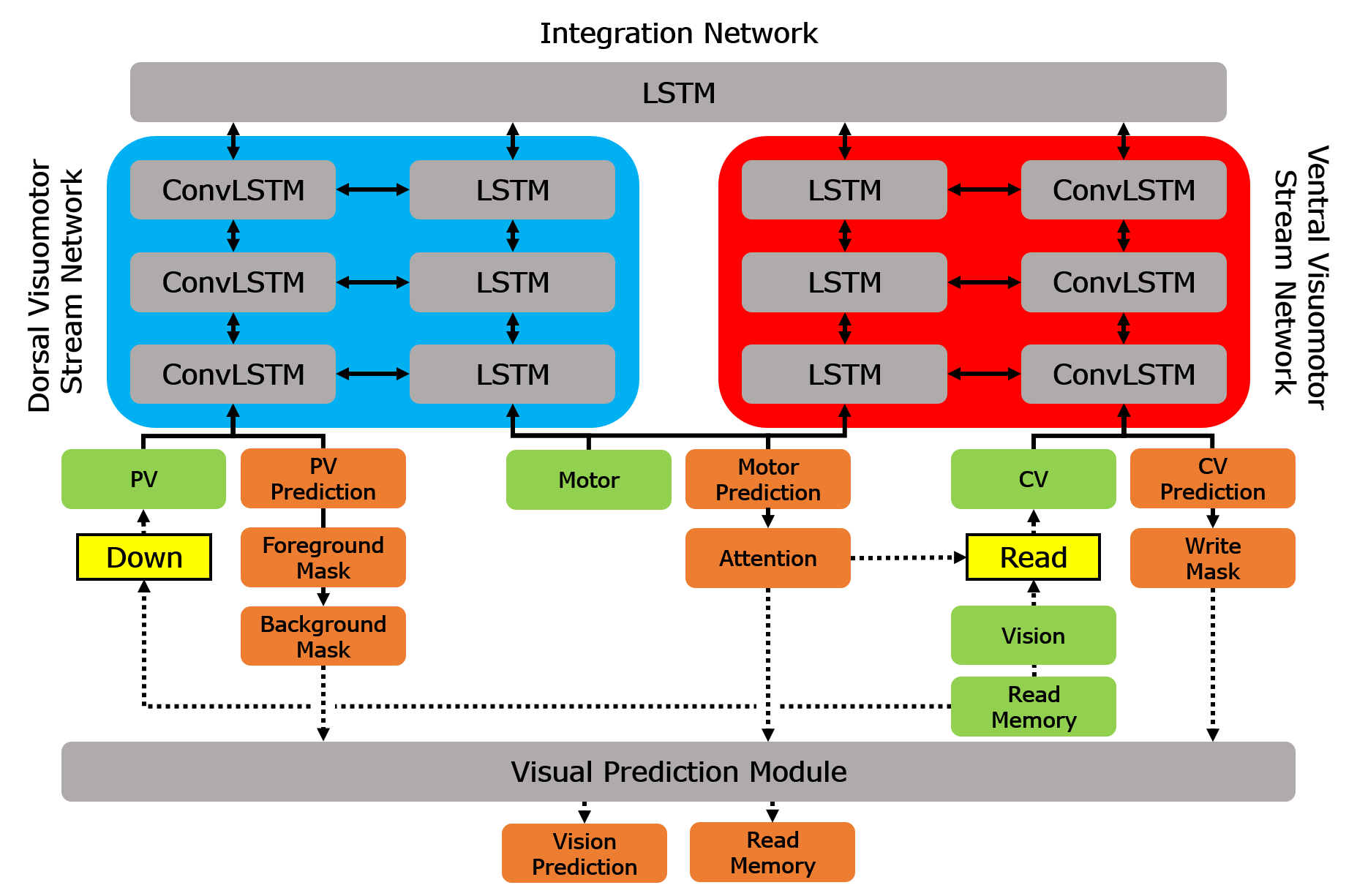}}
    \hfil   
    \subfloat[]{\includegraphics[width=0.4\textwidth]{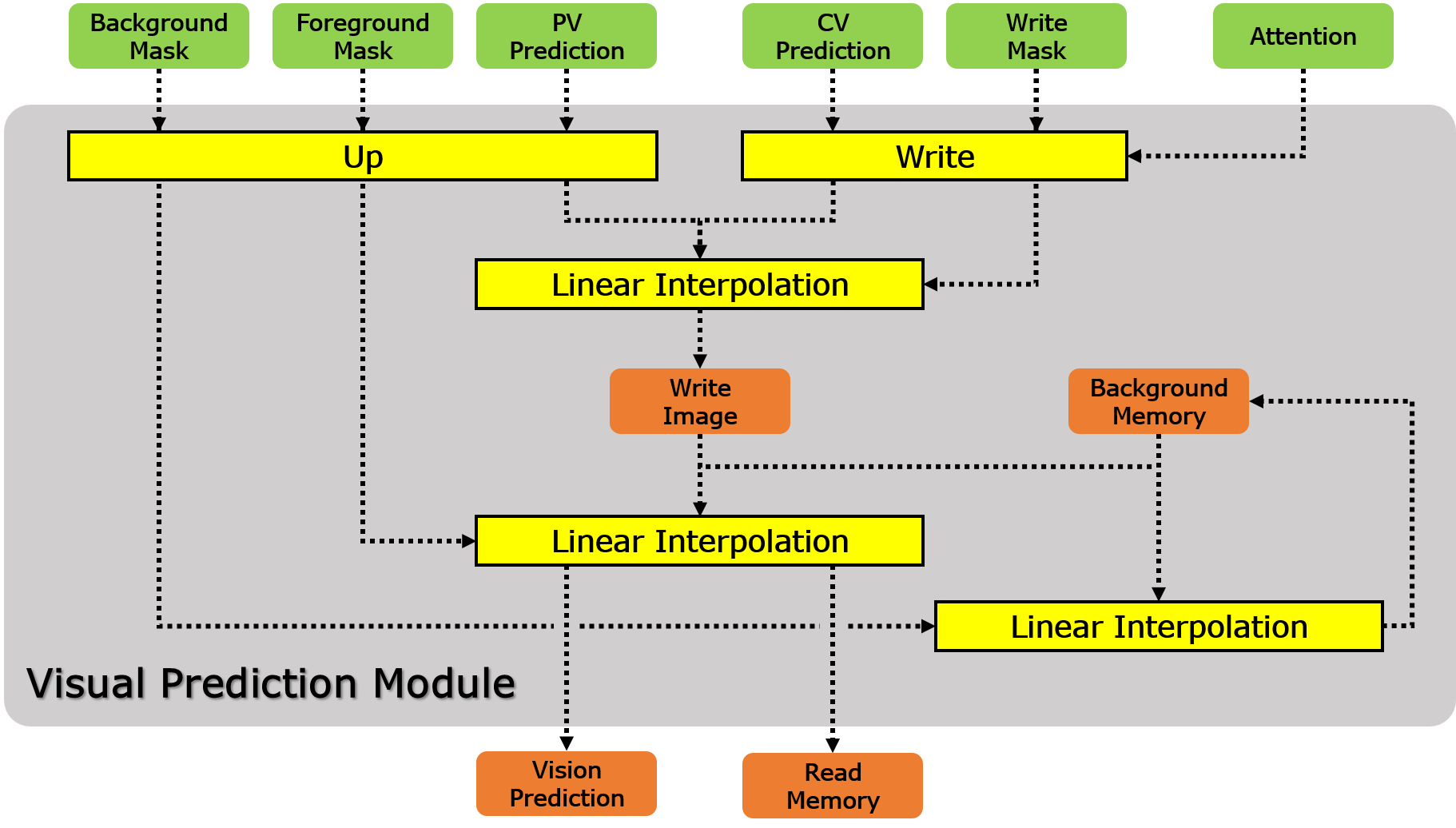}}
%\hfil
\caption{Our proposed model for goal-directed, long-term planning. (a) Overall model architecture. (b) Visual prediction module.}
%(c) Comparison between the baseline and AD when the update gate bias is initialized to -2.
\label{fig:DVM}
\vspace{-6mm}
\end{figure}

% \begin{figure}[thpb]
%   \centering
%   \framebox{\parbox{3in}{We suggest that you use a text box to insert a graphic (which is ideally a 300 dpi TIFF or EPS file, with all fonts embedded) because, in an document, this method is somewhat more stable than directly inserting a picture.
% }}
%   %\includegraphics[scale=1.0]{figurefile}
%   \caption{Inductance of oscillation winding on amorphous
%   magnetic core versus DC bias magnetic field}
%   \label{figurelabel}
% \end{figure}
We extend a predictive coding type deep visuomotor recurrent neural network model (P-DVMRNN) proposed in \cite{Choi:2018}. The proposed neural network model consists of a visual prediction module and three sub-networks: dorsal visuomotor stream, ventral visuomotor stream, and integration networks (Fig. \ref{fig:DVM}). 
% The overall architecture is similar to P-DVMRNN, but the main difference is that the proposed network has dual visuomotor streams rather than one. 
Inspired by the two stream hypothesis proposed in \cite{Norman:2002,Sheth:2016}, the dorsal visuomotor stream network receives peripheral visual information and the ventral visuomotor stream network receives central visual information, but both receive the same motor input. 
The size and location of the attentional focus for central vision vary continuously over time in response to changing environments and goals.
% Human attention is intrinsically dynamic, with focus continuously shifting
% Main assumption of this chapter is that segregation into the dorsal and ventral visuomotor streams based on the source of visual input (central and peripheral vision) will be help to generate an accurate sequence of actions. 
Each visuomotor stream network consists of visual and motor pathways. Each pathway has multiple recurrent layers with bottom-up and top-down connections between successive layers in the same pathway and a lateral connection between layers from the visual and motor pathways at the same level. 
The integration network integrates visuomotor information from two visuomotor streams.
Furthermore, the visual prediction module seamlessly merges peripheral and central visual predictions into a single visual prediction with background memory that preserves visuospatial information to predict future action outcomes.

\subsection{Dorsal Visuomotor Stream}
The dorsal visuomotor stream network consists of visual and motor pathways. For the visual pathway, we employ the network having several stacked convolutional long short-term memory (ConvLSTM) \cite{Shi:2015} layers to extract spatio-temporal features at multiple scales within a video. Each layer receives inputs from three sources: bottom-up, lateral, and top-down connections. 
The bottom-up connection carries information from an environment through vision to update internal states of the network and the top-down connection propagates the prediction or belief of an environment from higher to lower layers.
The lateral connection brings information from the motor pathway at the same level to enhance the interaction between visual and motor pathways. 
The convolution operation with stride is used for the bottom-up connection in order to reduce the size of feature maps. The lateral and top-down connections utilize the deconvolution or transposed convolution operation for expanding the size of the feature maps.
The bottom layer receives peripheral visual information $\mathbf{v}^{0}_{d, t}$ downsampled from the original image $\mathbf{v}^{0}_{t}$ and read memory $\mathbf{r}_{t}$ at each time step $t$. 
% By providing a downsampled image to the dorsal visuomotor network, ??.
The neural activations $\mathbf{v}^{l}_{d, t}$ in $l$th layer at each time step $t$ are computed as follows: 
\begin{equation}
\label{read}
    \mathbf{v}^{0}_{d, t} = downsample([\mathbf{v}^{0}_{t}, \mathbf{r}_{t}])
\end{equation}
\begin{equation}
\label{pv_vision}
    \mathbf{v}^{l}_{d, t} = 
    \begin{cases}
        ConvLSTM(\mathbf{v}^{l-1}_{d, t}, \mathbf{m}^{l}_{d, t-1},  \mathbf{a}_{t-1}),& \text{if } l= L\\
        ConvLSTM(\mathbf{v}^{l-1}_{d, t}, \mathbf{m}^{l}_{d, t-1}, \mathbf{v}^{l+1}_{d, t-1}),              & \text{otherwise}
    \end{cases}
\end{equation}
where $[\cdot]$ represent the concatenation operation along the channel dimension; $ConvLSTM(\cdot)$ represents the convolutional LSTM layer receiving three parameters that correspond to bottom-up, lateral, and top-down inputs, respectively; $\mathbf{m}^{l}_{d, t-1}$ and $\mathbf{a}_{t-1}$ represent neural activations of the motor pathway network in $l$th layer and the integration network at the previous time step, respectively.  $L$ represents the number of layers of the visuomotor stream network. 
%The top-down and lateral from motor expanded by using deconvolution operation.

At each time step $t$, the visual pathway network generates the prediction of peripheral vision $\mathbf{pv}_{t+1}$ for the next time step, the foreground mask $\mathbf{g}_{f, t+1}$, and the background mask $\mathbf{g}_{b, t+1}$ as follows: 
\begin{equation}
\label{pv_prediction}
    \mathbf{pv}_{t+1} = \tanh(deconv(\mathbf{v}^{1}_{d, t}))
\end{equation}
\begin{equation}
\label{foreground_mask}
    \mathbf{g}_{f, t+1} = \sigma(deconv(\mathbf{v}^{1}_{d, t}))
\end{equation}
\begin{equation}
\label{background_mask}
    \mathbf{g}_{b, t+1} = \sigma(deconv(\mathbf{v}^{1}_{d, t}))
\end{equation}
where $deconv(\cdot)$ represents the deconvolution or transposed convolution operation.

For the motor pathway, the network having stacked LSTM \cite{Hochreiter:1997} layers is used. As in the visual pathway network, each LSTM layer has bottom-up, lateral, and top-down connections. The neural activations $\mathbf{m}^{l}_{d, t}$ in $l$th layer at each time step $t$ are computed as follows: 
\begin{equation}
\label{pv_motor}
    \mathbf{m}^{l}_{d,t} =     
    \begin{cases}
        LSTM(\mathbf{m}^{l-1}_{d, t}, \mathbf{v}^{l}_{d, t-1},  \mathbf{a}_{t-1}),& \text{if } l= L\\
        LSTM(\mathbf{m}^{l-1}_{d, t}, \mathbf{v}^{l}_{d, t-1}, \mathbf{m}^{l+1}_{d, t-1}),              & \text{otherwise}
    \end{cases}
\end{equation}

\subsection{Ventral Visuomotor Stream}
The architecture of the ventral visuomotor stream network is same as the dorsal one. However, the crucial difference between two visuomotor stream networks is that the ventral network receives central visual information rather than peripheral visual information. For reading and writing central visual information, a spatial transformer network (STN) \cite{Jaderberg:2015} is used. STN is a module that generates attention parameters for cropping, translating, and scaling of an image. 
% By using a parameterized sampling kernel, STN is fully differentiable. This is important for the proposed model because the prediction error should be back-propagated for planning.
The bottom layer receives central visual information $\mathbf{v}^{0}_{v, t}$ attended from the original image $\mathbf{v}^{0}_{t}$ and read memory $\mathbf{r}_{t}$ by using STN with the attention parameters $\boldsymbol{\alpha}_{t}$ at each time step $t$.
The neural activations of the visual pathway $\mathbf{v}^{l}_{v, t}$ and the motor pathway $\mathbf{m}^{l}_{v, t}$ in $l$th layer at each time step $t$ are computed as follows: 
\begin{equation}
\label{cv_read}
    \mathbf{v}^{0}_{v, t} = read([\mathbf{v}^{0}_{t}, \mathbf{r}_{t}], \boldsymbol{\alpha}_{t})
\end{equation}
\begin{equation}
\label{cv_vision}
    \mathbf{v}^{l}_{v, t} = 
    \begin{cases}
        ConvLSTM(\mathbf{v}^{l-1}_{v, t}, \mathbf{m}^{l}_{v, t-1},  \mathbf{a}_{t-1}),& \text{if } l= L\\
        ConvLSTM(\mathbf{v}^{l-1}_{v, t}, \mathbf{m}^{l}_{v, t-1}, \mathbf{v}^{l+1}_{v, t-1}),              & \text{otherwise}
    \end{cases}
\end{equation}
\begin{equation}
\label{cv_motor}
    \mathbf{m}^{l}_{v,t} =     
    \begin{cases}
        LSTM(\mathbf{m}^{l-1}_{v, t}, \mathbf{v}^{l}_{v, t-1},  \mathbf{a}_{t-1}),& \text{if } l= L\\
        LSTM(\mathbf{m}^{l-1}_{v, t}, \mathbf{v}^{l}_{v, t-1}, \mathbf{m}^{l+1}_{v, t-1}),              & \text{otherwise}
    \end{cases}
\end{equation}
where $read(\cdot)$ represents the function to focus the sub-region of a given image by using STN with the attention parameters $\boldsymbol{\alpha}_{t}$ consisting of $s$, $t_x$, and $t_y$ for scaling, x-translation, and y-translation, respectively.

At each time step $t$, the visual pathway network generates the prediction of central vision $\mathbf{cv}_{t+1}$ for the next time step and the write mask $\mathbf{g}_{w, t+1}$ as follows:
\begin{equation}
\label{cv_prediction}
    \mathbf{cv}_{t+1} = \tanh(deconv(\mathbf{v}^{1}_{v, t}))
\end{equation}
\begin{equation}
\label{write_mask}
    \mathbf{g}_{w, t+1} = \sigma(deconv(\mathbf{v}^{1}_{v, t}))
\end{equation}

\subsection{Dual Visuomotor Stream Integration}
The integration network, having a single LSTM layer, is placed on top of the two visuomotor stream networks. Outputs of two visuomotor stream networks in the top layer are fed into the integration network. Hence, the integration network can integrate information from different modalities (vision and motor) and from different scales (peripheral and central vision). Integrated information containing the abstraction of an environment and the history of conducted actions. This abstract information in the integration network is projected to two visuomotor stream networks in order to predict the next visuomotor inputs and to interact with those networks.
The neural activations $\mathbf{a}_{t}$ at each time step $t$ are computed as follows: 
\begin{equation}
\label{integration}
    \mathbf{a}_{t} = LSTM(\mathbf{v}^{L}_{d, t}, \mathbf{v}^{L}_{v, t}, \mathbf{m}^{L}_{d, t}, \mathbf{m}^{L}_{v, t})
\end{equation}

\subsection{Visuomotor Generation}
Since the proposed model has two visuomotor stream networks, these need to be merged for visuomotor generation. The neural activations of dorsal and ventral motor pathways are used to generate the attention parameters $\boldsymbol{\alpha}_{t+1}$ and motor prediction $\mathbf{m}_{t+1}$ for the next time step as follows:
\begin{equation}
\label{eq:attention}
    \boldsymbol{\alpha}_{t+1} = MLP(\mathbf{m}^{1}_{d, t}, \mathbf{m}^{1}_{v, t})
\end{equation}
\begin{equation}
\label{eq:motor}
    \mathbf{m}_{t+1} = MLP(\mathbf{m}^{1}_{d, t}, \mathbf{m}^{1}_{v, t})
\end{equation}
where $MLP(\cdot)$ represents the multi-layer perceptron.

To generate a visual prediction $\mathbf{v}_{t+1}$ at the next time step, we propose a visual prediction module having three stages (Fig. \ref{fig:DVM}(b)). First, the peripheral visual prediction $\mathbf{pv}_{t+1}$ and central visual prediction $\mathbf{cv}_{t+1}$ from the dorsal and ventral visuomotor stream networks, respectively, are merged into a single image, called a write image, $\mathbf{w}_{t+1}$. Since the resolutions of peripheral and central vision are lower than of the original vision, the peripheral visual prediction $\mathbf{pv}_{t+1}$, foreground mask $\mathbf{g}_{f, t+1}$, and background mask $\mathbf{g}_{b, t+1}$ are upsampled; and the central visual prediction $\mathbf{cv}_{t+1}$ and write mask $\mathbf{g}_{w, t+1}$ are transformed to the original image using the attention parameters $\boldsymbol{\alpha}_{t+1}$ before merging. Note that the attention parameters $\boldsymbol{\alpha}_{t+1}$ are shared for writing and reading as in \cite{Eslami:2016}. Hence, the attention parameters $\boldsymbol{\alpha}_{t+1}$ will be used for reading the next image $\mathbf{v}^{0}_{t+1}$ as shown in Eq. (\ref{cv_read}). For smooth blending, the write mask $\mathbf{g}_{w, t+1}$ is used for the linear interpolation between the peripheral and central visual predictions.
Second, the visual prediction $\mathbf{v}_{t+1}$ is computed by interpolation between the write image $\mathbf{w}_{t+1}$ and background memory $\mathbf{bg}_{t}$ weighed by the foreground mask $\mathbf{g}_{f, t+1}$. The foreground masked background memory, called read memory $\mathbf{r}_{t+1}$, used for the visual prediction is given as a visual input for next time step. The background memory is important to preserve the central visual predictions at high resolution over time. The central visual prediction at the current time step can be overwritten at the next time step by occlusion, but it needs to be restored when occlusion is resolved. Finally, the next background memory $\mathbf{bg}_{t+1}$ is updated by the current background memory $\mathbf{bg}_{t}$ and write image $\mathbf{w}_{t+1}$ with the background mask $\mathbf{g}_{b, t+1}$.
All three stages explained above are computed as follows:
\begin{equation}
\label{write_image}
\begin{split}
    \mathbf{w}_{t+1}  ={} & (1 - f_{w}(\mathbf{g}_{w, t+1},  \boldsymbol{\alpha}_{t+1})) \odot f_{up}(\mathbf{pv}_{t+1})  \\
    & + f_{w}(\mathbf{g}_{w, t+1},  \boldsymbol{\alpha}_{t+1}) \odot f_{w}( \mathbf{cv}_{t+1},  \boldsymbol{\alpha}_{t+1})
\end{split}
\end{equation}
% \begin{equation}
% \label{integration}
% \begin{split}
%     \mathbf{w}_{t+1}  ={} & (1 - write(\mathbf{g}_{w, t+1},  \boldsymbol{\alpha}_{t+1})) \odot upsample(\mathbf{pv}_{t+1})  \\
%     & + write(\mathbf{g}_{w, t+1},  \boldsymbol{\alpha}_{t+1}) \odot write( \mathbf{cv}_{t+1},  \boldsymbol{\alpha}_{t+1})
% \end{split}
% \end{equation}
\begin{equation}
\label{read_memory}
    \mathbf{r}_{t+1} = (1 - f_{up}(\mathbf{g}_{f, t+1})) \odot \mathbf{bg}_{t} 
\end{equation}
\begin{equation}
\label{vision_integration}
    \mathbf{v}_{t+1} = \mathbf{r}_{t+1} + f_{up}(\mathbf{g}_{f,t+1}) \odot \mathbf{w}_{t+1}
\end{equation}
\begin{equation}
\label{background}
    \mathbf{bg}_{t+1} = (1 - f_{up}(\mathbf{g}_{b, t+1})) \odot \mathbf{bg}_{t} + f_{up}(\mathbf{g}_{b,t+1}) \odot \mathbf{w}_{t+1}
\end{equation}
% \begin{equation}
% \label{background}
% \begin{split}
%     \mathbf{bg}_{t+1} = {} & (1 - upsample(\mathbf{g}_{b, t+1})) \odot \mathbf{bg}_{t} \\
%     & + upsample(\mathbf{g}_{b,t+1}) \odot \mathbf{w}_{t+1}
% \end{split}
% \end{equation}
where $\odot$ represents the element-wise multiplication, $f_{up}(\cdot)$ represents the upsampling function, and $f_{w}(\cdot)$ represents the write function for transforming an attended sub-region to an original image.

\subsection{Sampling Initial States Based on Variational Bayes}
Initial sensitivity is one of important characteristics of recurrent neural network (RNN) models \cite{Nishimoto:2004}. The proposed model is trained with the mapping between the training visuomotor sequences and corresponding initial states. Once the training is finished successfully, the model is able to generate multiple visuomotor sequences based solely upon the corresponding trained initial states without any external input. Hence, the trained initial state can be considered an intention, which should be prepared before executing goal-directed actions, as in the brain \cite{Shima:2006}. Most previous models utilizing initial sensitivity were trained using a deterministic approach \cite{Nishimoto:2004,Yamashita:2008,Choi:2018}.
However, deterministic models have difficulty dealing with stochasticity and uncertainty in an environment, leading to blurry predictions \cite{Kingma:2013,Babaeizadeh:2017,Denton:2018}.
Hence, in this paper, we use a variational Bayes approach \cite{Kingma:2013} for training the initial states in order to capture the full distribution of outcomes. The initial state $\mathbf{s}^{n}_{0}$ for the $n$th training sample is sampled using the reparametrization trick \cite{Kingma:2013} as follows:
\begin{equation}
\label{eq:reparameterization}
    \mathbf{s}^{n}_{0} = \boldsymbol{\mu}^{n}_{0} + \boldsymbol{\epsilon} \odot \boldsymbol{\sigma}^{n}_{0} %\sim \mathcal{N}(\boldsymbol{\mu}^{n}_{0}, \boldsymbol{\sigma}^{n}_{0})
\end{equation}
\begin{equation}
\label{eq:sampling}
    \boldsymbol{\epsilon} \sim \mathcal{N}(\mathbf{0}, \mathbf{I})
\end{equation}
where $\boldsymbol{\mu}^{n}_{0}$ and $\boldsymbol{\sigma}^{n}_{0}$ represent the mean and standard deviation of the initial state $\mathbf{s}^{n}_{0}$, respectively, and $\boldsymbol{\epsilon}$ represents an auxiliary noise sampled from a standard normal distribution.

\subsection{Training}
The proposed model is trained to mentally generate multiple visuomotor sequences with corresponding initial states. There are two types of initial state: prior and posterior. The prior initial state is shared for all training samples, but posterior initial states are provided for each training sample. Each initial state is parameterized by the mean and standard deviation, which are updated directly during the training phase. Note that we train the prior as well as posterior initial states as in \cite{Denton:2018}. The weights, biases, and initial states of the model are updated to minimize both the visuomotor prediction error and the Kullback-Leibler divergence between prior and posterior initial states, which is same as maximizing the variational lower bound as in the variational auto-encoder (VAE) \cite{Kingma:2013}. The loss $L^n$ for the $n$th training sample is defined as follows:
\begin{equation}
\label{eq:loss_train}
    L^{n} = L^{n}_{v} + L^{n}_{m} + \beta D_{KL}(q_{\boldsymbol{\phi}^{n}}(\mathbf{s}^{n}_{0})||p_{\boldsymbol{\theta}}(\mathbf{s}_{0}) ) 
\end{equation}
\begin{equation}
\label{Loss:vision}
    L^{n}_{v}  = \sum_{t=1}^{T^{n}}L^{n}_{v, t}  = \sum_{t=1}^{T^{n}}(\mathbf{v}^{n}_{t} - \mathbf{\tilde{v}}^{n}_{t})^{2} + (f_{up}(\mathbf{pv}^{n}_{t}) - \mathbf{\tilde{v}}^{n}_{t})^{2} 
\end{equation}
\begin{equation}
\label{Loss:motor}
    L^{n}_{m} = \sum_{t=1}^{T^{n}}L^{n}_{m, t} = \sum_{t=1}^{T^{n}}(D_{KL}(\mathbf{m}^{n}_{t}||\mathbf{\tilde{m}}^{n}_{t}))
\end{equation}
where $\boldsymbol{\phi}^{n}$ and $\boldsymbol{\theta}$ represent parameters of the posterior initial state for the $n$th training sample and prior initial state, respectively; $D_{KL}(\cdot)$ represents the Kullback-Leibler divergence; $\beta$ represents the hyper-parameter for controlling the balance between minimizing the reconstruction error and fitting the prior \cite{Denton:2018}; $T^n$ represents the length of the $n$th target visuomotor sequence; and $\mathbf{\tilde{v}}^{n}_{t}$ and $\mathbf{\tilde{m}}^{n}_{t}$ represent the $n$th target visual and motor values at each time step $t$. The loss $L^{n}$ is for the case of a single training sample, but it can be easily extended to the case of a mini-batch training samples.

During the training phase, we follow the closed-loop training method used in \cite{Yamashita:2008} to improve the mental simulation capability of the model. In the closed-loop training, the visual and motor predictions are fed into the model as input for the next time step. However, since it is hard to train the model using complete prediction feedback, external and predicted visuomotor values are blended as follows:
\begin{equation}
\mathbf{v}^{0}_{t} = 0.9  \mathbf{v}_{t} + 0.1 \mathbf{\tilde{v}}_{t}
\end{equation}
\begin{equation}
\mathbf{m}^{0}_{t} = 0.9  \mathbf{m}_{t} + 0.1 \mathbf{\tilde{m}}_{t}
\end{equation}

\subsection{Planning}
During the planning phase, the loss $L^n$ in Eq. (\ref{eq:loss_train}) is no longer available because only a few initial visuomotor values and a final image are given to the model. 
Therefore, the loss $L^{n}_{p}$ for planning the $n$th testing sample is defined as follows:
\begin{equation}
\label{eq:loss_plan}
    L^{n}_{p} = \sum_{t=1}^{T_{g}} (L^{n}_{v, t} + L^{n}_{m, t}) + L^{n}_{v, T_{e}} + \beta D_{KL}(q_{\boldsymbol{\psi}^{n}}(\mathbf{s}_{n, 0})||p_{\boldsymbol{\theta}}(\mathbf{s}_{0}) ) 
\end{equation}
where $\boldsymbol{\psi}^{n}$ represents parameters of the initial state for the $n$th testing sample, $T_{g}$ represents the number of initial target visuomotor values given to the model, and $T_{e}$ represents the end step for visuomotor generation.
Note that only initial states are updated to minimize the loss $L_{p}$ while the weights and biases of the model are fixed during the planning phase.

\section{Results}

\subsection{Implementation Details}
% Both dorsal and ventral visuomotor stream networks have the same architecture and parameters.
Each visual pathway network of both dorsal and ventral visuomotor stream networks consisted of three ConvLSTM layers having 16, 32, and 64 feature maps, respectively. The convolution kernel, stride, and padding sizes were set to 5$\times$5, 2$\times$2, and 2$\times$2, respectively, for bottom-up connections. The deconvolution kernel, stride, and padding sizes were set to 6$\times$6, 2$\times$2, and 2$\times$2, respectively, for top-down connections, except from the integration network. For lateral connections from the motor pathway network and the top-down connection from the integration network, the convolutional kernel size was set the same as the resolution of feature maps at each layer in the visual pathway network. The resolution of peripheral and central vision were set to half the original vision.
Each motor pathway network of both visuomotor stream networks consisted of three LSTM layers having 512, 256, and 128 neurons, respectively. To generate motor and attention at each time step as shown in Eq. (\ref{eq:attention}) and (\ref{eq:motor}), the multi-layer perceptron (MLP) with one hidden layer of 256 neurons, layer normalization (LN) \cite{Ba:2016}, and a rectified linear unit (ReLU) activation function was used.
The integration network consisted of a single LSTM layer having 512 neurons.
% The network configuration is summarized in Table \ref{}.

For training, Adam optmizer \cite{Kingma:2014} was applied to train the weights, biases, and initial states of the model over 1000 epochs. The learning rate of the weights, biases, and prior initial state was set to 5e-4. Since each posterior initial state is updated only when its corresponding training sample is given to the model, the learning rate of the posterior initial states was set to 0.015. Mini-batch size was set to 10. The hyper-parameter $\beta$ in Eq. (\ref{eq:loss_train}) and (\ref{eq:loss_plan}) was set to 1e-5.
An L2-norm weight decay of 1e-4 was applied while updating the model parameters to prevent over-fitting. Because of the exploding gradient problem, we employed a gradient clipping method \cite{Pascanu:2013} which rescales the L2-norm of the gradient to a threshold whenever the L2-norm exceeds that threshold. Here, the threshold was set to 0.2. Each mean and standard deviation of both prior and posterior initial states were initialized to 0 and 1. 

% All weights were initialized using randomly selected values from a zero-mean Gaussian distribution, with the standard deviation $\sigma$ set differently depending on the experiment. Similar to the bias initialization trick used to solve the gradient vanishing problem of LSTM \cite{Gers:2000,Jozefowicz:2015}, update gate biases were initialized to -2, and the remaining biases were initialized to 0 by default (unless otherwise noted). For both recurrent batch normalization (recurrent BN) \cite{Cooijmans:2017} and layer normalization (LN) \cite{Ba:2016}, the gain $\gamma$ and bias $\beta$ of each affine transformation were initialized to 1 and 0, respectively. However, when recurrent BN and LN were applied to the update gate, the bias $\beta$ of each affine transformation was initialized to -2 according to the above mentioned bias initialization trick. Note that we did not initialize the gain $\gamma$ to 0.1 as did Cooijmans et al. \cite{Cooijmans:2017} because initializing the gain $\gamma$ to 1 produced better results in the following experiments.

\subsection{Evaluation Protocol}
During the planning phase, the model inferred the posterior initial state of a testing sample by iteratively updating the initial state toward to minimize the loss $L_{p}$ defined in Eq. (\ref{eq:loss_plan}). $T_{g}$ and $T_{e}$ were set to 5 and 100. The prior initial state was used to initialize the initial state as a starting point of planning.
The error regression was repeated over 50 epochs.
At each epoch, 16 visuomotor sequences were generated based on 16 initial states sampled by using the reparameterization trick in Eq. (\ref{eq:reparameterization}) with the mean and standard deviation of the current initial state, and then the loss $L_{p}$ was computed for each sequence. The sequence having the minimum loss $L_{p}$ over 16 visuomotor sequences was used for updating the initial state.
We repeated the above error regression procedure 6 times and selected the visuomotor sequence having the minimum loss $L_{p}$ as an optimal plan.

\subsection{RGB Block Stacking}
\subsubsection{Experiment Setting}
% \begin{figure}[!t]
% \centering
%     \subfloat[]{\includegraphics[width=0.1\textwidth]{fig/exp/RGB_stack_init_config.png}
%     \label{fig:}}
%     \subfloat[]{\includegraphics[width=0.1\textwidth]{fig/exp/RGB_stack_final_config_1.png}
%     \label{fig:}}
%     % \hfil 
%     \subfloat[]{\includegraphics[width=0.1\textwidth]{fig/exp/RGB_stack_final_config_2.png}
%     \label{fig:}}
%     \subfloat[]{\includegraphics[width=0.1\textwidth]{fig/exp/RGB_stack_final_config_3.png}
%     \label{fig:}}
% %\hfil
% \caption{Sample images for (a) initial block configuration and (b)-(d) final block configurations, each given as a goal.}
% %(c) Comparison between the baseline and AD when the update gate bias is initialized to -2.
% \label{fig:RGB_stacking_example}
% \vspace{-6mm}
% \end{figure}

We conducted a robot manipulation experiment using a Torobo Arm robot manufactured by Tokyo Robotics. Torobo Arm has a total of 8 joints including the end effector, of which 6 joints were used in the following experiments. Each motor joint was converted to a 10 dimensional sparsely encoded vector as in \cite{Yamashita:2008} to reduce the overlap between motor sequences caused by the low dimensionality of the motor. The task space for the robot was approximately 30$cm^2$, and an RGB camera was fixed facing both the task space and the robot. While collecting data, the robot joint angles and video frames were sampled at 20Hz. All data were temporally downsampled 7 times to reduce computational and memory costs. All video frames were rescaled to 64$\times$64 pixels and normalized to -1 to 1.

For the experimental task, we used three colored cubes (red (R), blue (B), and green (G)) size of 5$cm^3$. 
% Data collection was automated by first randomly sampling a location for each block from a $n\times n$ grid, then having the robot initialize the task by placing each block at the corresponding location (Fig. \ref{fig:RGB_stacking_example}), return to the home position, and then finally execute the stacking operation. 
Data collection was automated according to the following steps:
(1) randomly sampling a location for each block from a $n\times n$ grid, (2) having the robot initialize the task by placing each block at the corresponding location, (3) returning the robot to the home position, and (4) executing the stacking operation. 
The block location grid size $n$ was set to 10 for training and 8 for testing to check the generalization capability of the model with an unlearned location distribution. For the stacking operation, we sampled three possible combinations of blocks as goal states, which the robot should achieve. Depending on the sampled goal state, the robot generated the corresponding motor trajectory while recording both video and motor sequence. There are total six possible combinations for stacking three color blocks, but we allowed only four stacking combinations (RGB, RBG, GRB, and GBR) for training by excluding remaining two stacking combinations (BRG and BGR) when the blue block be the base block. During the testing phase, these two unlearned stacking combinations were used to evaluate the generalization capability of the model. We used 300 videos for training and 45 videos for testing.
The objective of the experiment is whether the model can generate correct action planning or not when a goal image is given. The task is difficult to achieve because the model needs to recognize what is the goal state based on a given goal image and then generates a correct visuomotor sequence for completing the goal. 

\subsubsection{Visual Prediction Mechanism}
% \begin{figure*}[!t]

% \centering
%     \subfloat[]{\includegraphics[width=0.8\textwidth]{fig/exp/visual_prediction_module/pv.png}
%     \label{fig:}}
%     \hfil
%     \subfloat[]{\includegraphics[width=0.8\textwidth]{fig/exp/visual_prediction_module/cv.png}
%     \label{fig:}}
%     \hfil
%     \subfloat[]{\includegraphics[width=0.8\textwidth]{fig/exp/visual_prediction_module/bg.png}
%     \label{fig:}}
%     \hfil
%     % \hfil
%     \subfloat[]{\includegraphics[width=0.8\textwidth]{fig/exp/visual_prediction_module/rv.png}
%     \label{fig:}}
%     % \hfil
%     % \subfloat[]{\includegraphics[width=0.2\textwidth]{fig/exp/visual_prediction_module/tv.png}
%     % \label{fig:}}
    
% %\hfil

% \caption{Visual mental simulation. (a) Peripheral visual prediction. (b) Central visual prediction. (c) Background memory. (d) Raw visual prediction.} %All figures are temporally downsampled four times for visualization.}
% \label{fig:vision_prediction}
% \end{figure*}

\begin{figure*}[!t]

\centering
\includegraphics[width=\textwidth]{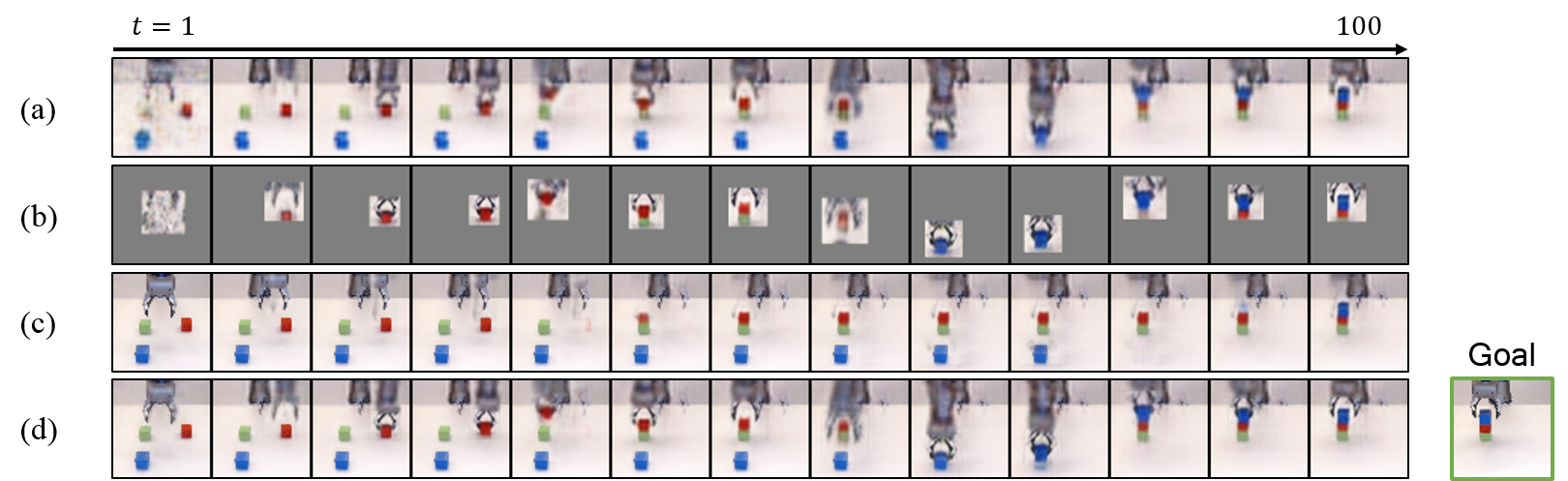}
%\vspace{-3mm}
\caption{Example of visual mental simulation. (a) Peripheral visual prediction. (b) Central visual prediction. (c) Background memory. (d) Visual prediction.} %All figures are temporally downsampled four times for visualization.}
\label{fig:vision_prediction}
\vspace{-6mm}
\end{figure*}
% Since a goal is given as an image, the capability of visual mental simulation is . 
% However, the visual prediction mechanism of the proposed model is com
% Due to the segregation of peripheral and central vision and the background memory, the visual prediction mechanism of the proposed model is much more complex than P-DVMRNN \cite{Choi:2018}. %motor prediction, which is one of the main contributions of this paper. 
Due to the segregation of peripheral and central vision and the background memory, the visual prediction mechanism of the proposed model is much more complex than previous unstructured predictive models \cite{Finn:2017,Choi:2018}.
Therefore, we qualitatively analyzed how the proposed model mentally generated visual sequences after planning was finished (Fig. \ref{fig:vision_prediction}). 
% Fig. \ref{fig:vision_prediction}(a) and (b) show the peripheral and central visual predictions from the dorsal and ventral visual pathways, respectively. 
The peripheral visual prediction covered the full receptive field, but its quality was low due to the resolution limitation. On the other hand, the quality of central visual prediction was high by focusing on a narrow receptive field, called an attention window. By utilizing a dynamic top-down visual attention, central vision maximally sampled visual information relevant to the given goal. Interestingly, we observed that the attention window tracked the robot end effector without any supervision. 
A possible interpretation can be made in terms of a retinal slip, which is created by the difference between eye velocity and the motion of a target object. 
During tracking motions, the eye accelerates in the direction of target motion to reduce the retinal slip. 
As in previous work \cite{Wijesinghe:2018}, the proposed model tried to locate the attention window on the end effector to minimize the vision loss, which can be interpreted as the retinal slip caused by the movement of the end effector. 
This is analogous to smooth pursuit eye movements in primates.
% For another interpretation, the end effector movement generates a retinal slip, which is motion of a visual image on the retina. 
% The visual prediction error caused by the retinal slip is hard to minimize by using the peripheral visual prediction, so that 
% Hence, the attention window tracks the end effector in order to cancel the retinal slip 
% However, note that attention is generated by motor pathways.

In order to effectively predict high-resolution visual sequences, the visual prediction module mentally generates the visual prediction by merging the peripheral and central visual predictions, which complement each other with regard to resolution and the size of a receptive field, with the background memory. 
% However, it is not enough to long-term visual prediction because central visual prediction. 
% Since the peripheral and central visual predictions from the dorsal and ventral visuomotor streams, respectively, are complementary each other, two visual predictions should be merged into the visual prediction.
% The visual prediction module generates the visual prediction by combining the background memory as well as the peripheral and central visual predictions. 
Background memory is a core part of the visual prediction module for long-term visuospatial information maintenance. As shown in Fig. \ref{fig:vision_prediction}(c), background memory focused the block configuration at each time step and ignored robot movement, which often caused block occlusion. With background memory, the model was able to store the information of the block configuration even when the block(s) were completely occluded by the robot and to use it for visuomotor prediction in the future. This demonstrates that background memory works similarly to visual working memory, based solely on learning.
% We will further investigate the importance of the background memory in later section. 
Finally, Fig. \ref{fig:vision_prediction}(d) shows the visual prediction generated by the visual prediction module. The visual prediction looks sharper than the peripheral visual prediction (especially for the gripper and blocks) because of the central visual prediction and background memory.

\subsubsection{Visuomotor Contingency}
One of the characteristics of the proposed model is that it predicts both visual and motor sequences jointly compared with previous work in which motor sequences were used as auxiliary inputs for video prediction \cite{Finn:2017,Babaeizadeh:2017,AXLee:2018}. 
Fig. \ref{fig:qualitative_comparision}(a) shows that the visual and motor predictions were highly correlated. 
It means that the proposed model learned the relationship between visual and motor sequences, called visuomotor contingency.
Therefore, visual prediction error minimization leads to motor prediction error minimization and vice versa, which is the reason that goal-directed visuomotor planning is possible only with a goal image. 
% Furthermore, note that the dorsal and ventral motor pathways generate the visual attention parameters as well as motor prediction at each time step.
% As like the motor prediction, the aspect of visual attention changes completely over epochs during the error regression (Fig. \ref{fig:planning} (d)-(f)).
% In brief, the proposed model simultaneously plans visual and motor sequences in addition to visual attention by using the error regression.

\subsubsection{Importance of Background Memory}

% \begin{figure*}[!t]
% \centering
%     \subfloat[]{\includegraphics[width=0.8\textwidth]{fig/exp/comparision/with_background/vision.png}
%     \label{fig:}}
%     \hfil
%     \subfloat[]{\includegraphics[width=0.8\textwidth]{fig/exp/comparision/without_background/vision.png}
%     \label{fig:}}
%     \hfil
%     \subfloat[]{\includegraphics[width=0.8\textwidth]{fig/exp/comparision/deterministic/vision.png}
%     \label{fig:}}
%     \hfil
%     \subfloat[]{\includegraphics[width=0.25\textwidth]{fig/exp/comparision/with_background/motor_6.pdf}
%     \label{fig:}}
%     \subfloat[]{\includegraphics[width=0.25\textwidth]{fig/exp/comparision/without_background/motor_6.pdf}
%     \label{fig:}}
%     \subfloat[]{\includegraphics[width=0.25\textwidth]{fig/exp/comparision/deterministic/motor_6.pdf}
%     \label{fig:}}
% %\hfil
% \caption{Qualitative comparison of planning performance for RBG stacking. (a)-(c) Visual prediction. (d)-(f) Motor prediction. Solid and dashed lines represent the predicted and target motor sequences, respectively. Each column corresponds to the proposed network, the network without the background memory, and the deterministic network (from left to right).}
% \label{fig:qualitative_comparision}
% \end{figure*}

\begin{figure*}[!t]
\centering
\includegraphics[width=\textwidth]{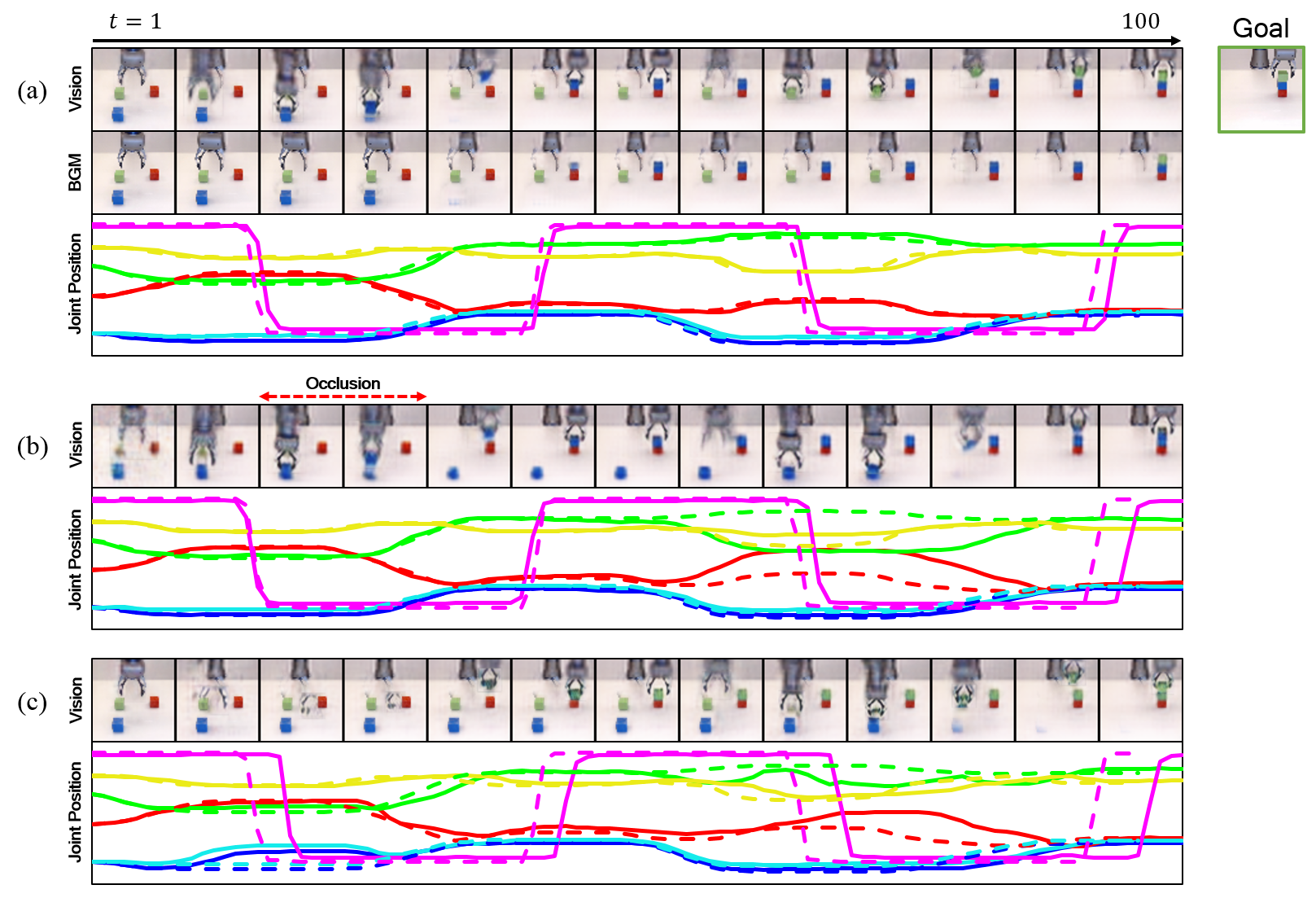}
\vspace{-7mm}
%\hfil
\caption{Qualitative comparison of planning performance. Solid and dashed lines represent the predicted and target motor sequences, respectively. (a) Proposed model. (b) Model without background memory. Without background memory, the model forgets the green block during occlusion (red dotted arrow). (c) Deterministic model. The model fails to find an adequate plan due to over-fitting. Abbreviation: BGM, background memory.}
\label{fig:qualitative_comparision}
\vspace{-6mm}
\end{figure*}

% \begin{figure*}[!t]
% \centering
%     \subfloat[]{\includegraphics[width=0.2\textwidth]{fig/exp/comparision/with/vision_6.png}
%     \label{fig:}}
%     \subfloat[]{\includegraphics[width=0.2\textwidth]{fig/exp/comparision/with/vision_32.png}
%     \label{fig:}}
%     \subfloat[]{\includegraphics[width=0.2\textwidth]{fig/exp/comparision/with/vision_44.png}
%     \label{fig:}}
%     \hfil
%     \subfloat[]{\includegraphics[width=0.2\textwidth]{fig/exp/comparision/with/background_6.png}
%     \label{fig:}}
%     \subfloat[]{\includegraphics[width=0.2\textwidth]{fig/exp/comparision/with/background_32.png}
%     \label{fig:}}
%     \subfloat[]{\includegraphics[width=0.2\textwidth]{fig/exp/comparision/with/background_44.png}
%     \label{fig:}}
% %\hfil
% \caption{Importance of background memory. (a)-(c) Visual prediction. (d)-(f) Background memory. Each column represents three selected testing samples.}
% \label{fig:qualitative_background}
% \end{figure*}

\begin{figure}[!t]
\centering
    \subfloat[]{\includegraphics[width=0.24\textwidth]{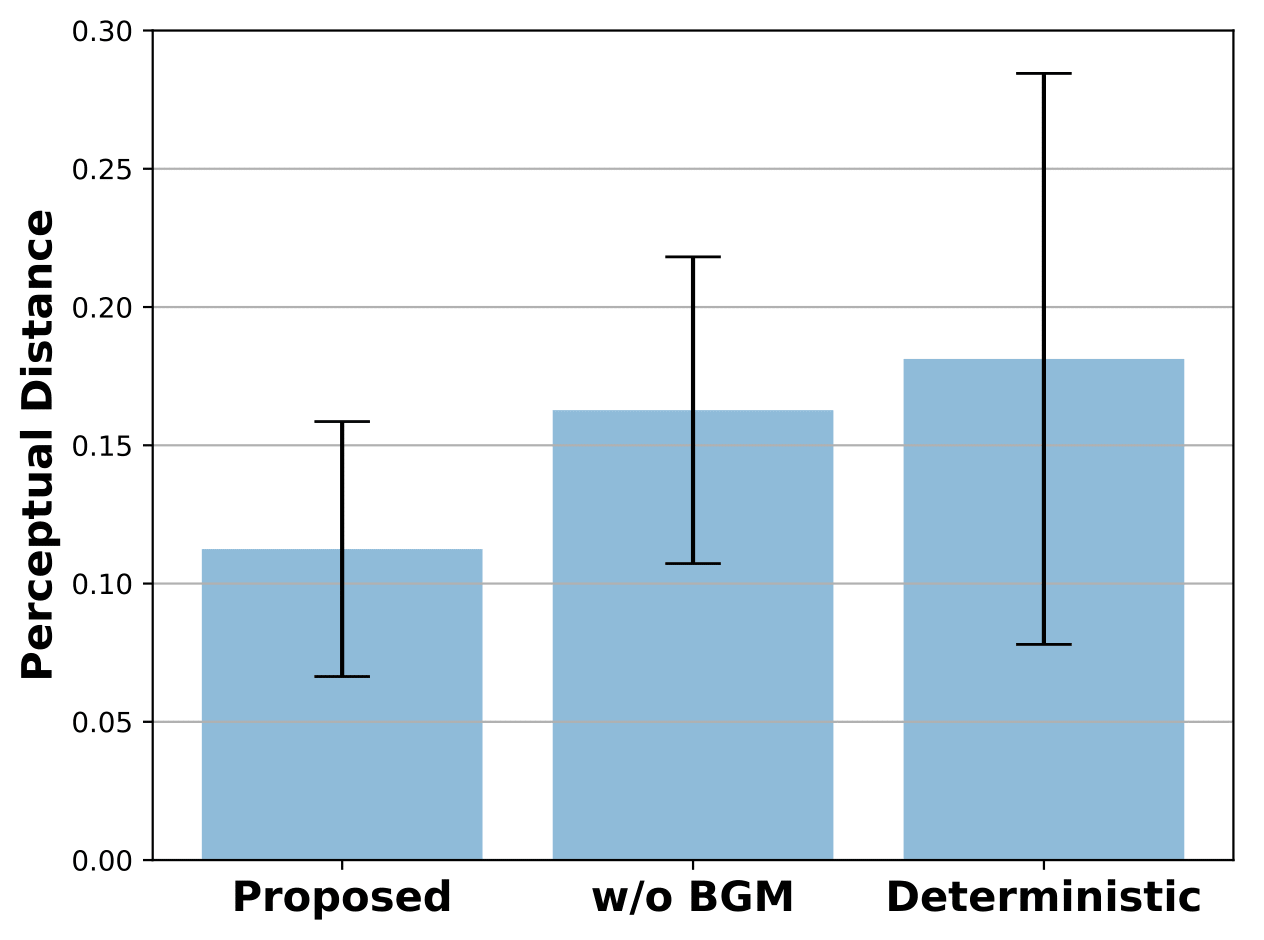}}
    %\hfil
    \subfloat[]{\includegraphics[width=0.24\textwidth]{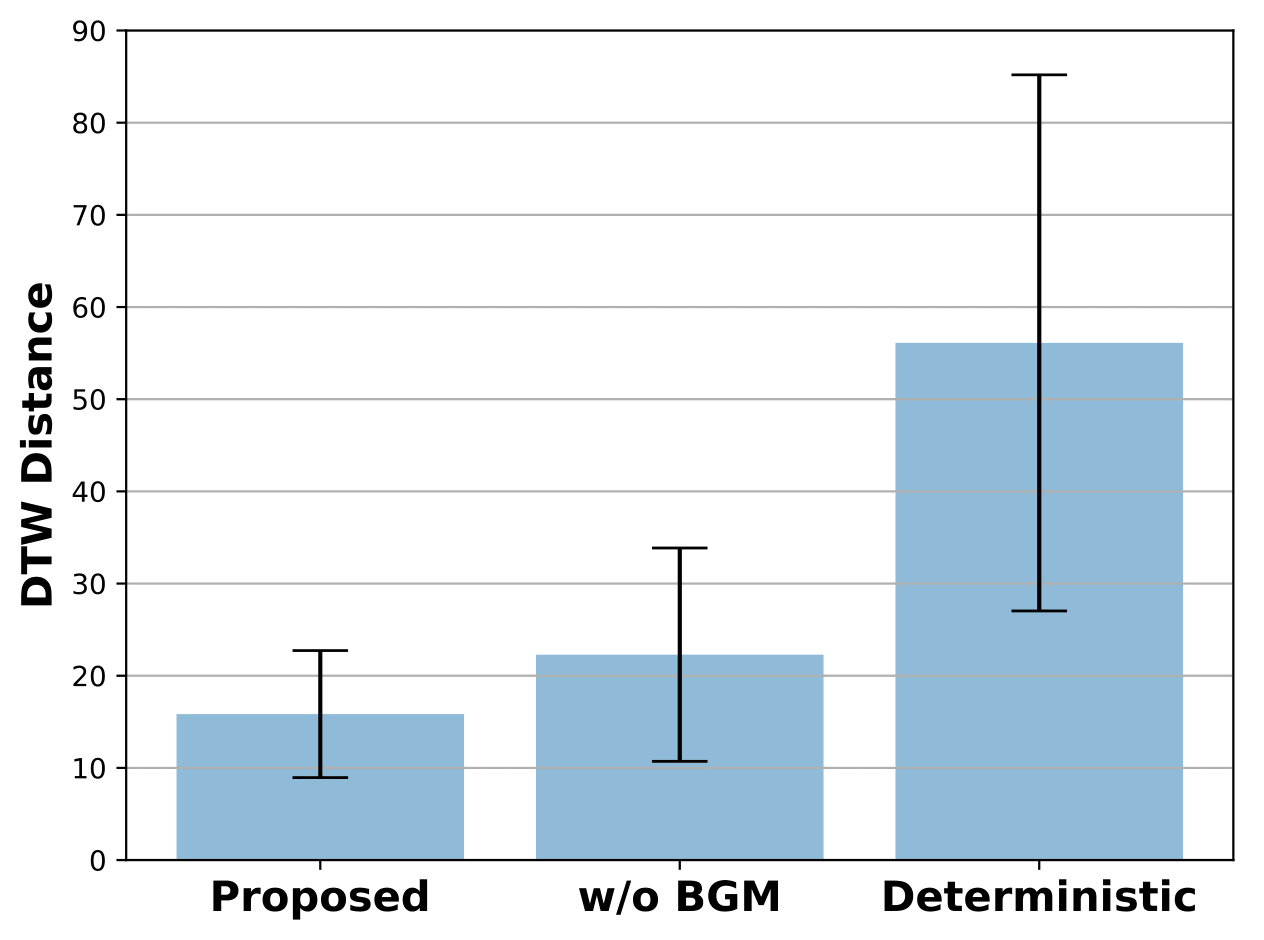}
    }
%\hfil
\caption{Quantitative comparison of planning performance. (a) Visual prediction evaluation. (b) Motor prediction evaluation. Bars and error bars represent the means and standard deviations, respectively.}
\label{fig:quantitative_comparision}
\vspace{-6mm}
\end{figure}

To investigate the importance of background memory for visuomotor planning, we compared the planning performance between the proposed model and its variant without background memory in a qualitative and quantitative manner.
In the model without background memory, background memory was removed in the visual prediction module and read memory was not given to the model as an input.
% The model without the background memory was exactly the same as the proposed model except that the background memory was removed in the visual prediction module and the read memory was not given to the model as an input.
For quantitative evaluation, we used two different evaluation metrics: perceptual loss \cite{Zhang:2018} and dynamic time warping (DTW) \cite{Sakoe:1978} for visual and motor prediction evalution, respectively.
It has been reported that conventional pixel-wise metrics for video prediction, such as PSNR and SSIM, are far from human perception because these assume pixel-level independence \cite{Babaeizadeh:2017, AXLee:2018}.
Hence, we used a perceptual metric proposed in \cite{Zhang:2018} as the evaluation metric of visual prediction. 
DTW \cite{Sakoe:1978} is used for the evaluation metric of motor prediction because the predicted motor sequence can be longer or shorter than the target motor sequence, which is unknown until it is finished.
DTW measures the similarity between two sequences, which may vary in time or speed.

We hypothesized that elimination of background memory will severely degrade the performance of planning because  background memory takes charge of visuospatial information maintenance. 
For a qualitative comparison between the proposed model and ablations of the proposed model, we illustrated the visuomotor sequences that were mentally generated by each models (Fig. \ref{fig:qualitative_comparision}).  
As we hypothesized, the model without  background memory failed to generate the green block after the robot picked the blue block due to occlusion.
To compensate for the green block that had disappeared, the model created an imaginary blue block where the blue block was originally located after the blue block was picked, and then picked this imaginary blue block for the second stacking.
Obviously, the corresponding motor prediction was different from the target motor sequence during the second half of mental simulation due to the repetition of the blue block picking and failed to achieve the desired goal.
On the other hand, in the proposed model, the visual prediction immediately recovered the green block after the occlusion was resolved and the motor prediction was well matched with the target motor sequence.
The reason is that the background memory preserves and updates the information of the block configuration, which is used by the visual prediction module for visual prediction of future actions. 
% Furthermore, the visual prediction motor prediction
% Three more samples support that the background memory is able to extract the block c The results in Fig. \ref{fig:qualitative_background} support that the background memory .
Fig. \ref{fig:quantitative_comparision} shows a quantitative comparison of the effect of background memory.
The proposed model consistently outperformed the model without background memory in terms of visual and motor predictions.
%%%%% ADD SENTENCE FUTURE %%%%%%%
Both qualitative and quantitative results demonstrated that background memory is essential for visuomotor planning.
%helps to predict better visual sequences and the clear block configuration motor prediction.  which is the information for visually guided object manipulation

% \subsubsection{Importance of Background Memory}
\subsubsection{Deterministic versus Stochastic Initial States}
As mentioned in \cite{Choi:2018}, deterministic predictive models require a huge number of training samples to achieve good generalization. 
However, it is difficult and time-consuming to prepare an enough training samples, especially for real robot experiments.
Furthermore, the required number of training samples exponentially increases as the complexities of tasks and models increase. 
To address this limitation, we proposed a stochastic predictive model under variational predictive coding because Bayesian approaches prevent over-fitting on few training samples \cite{Ghahramani:2015}.
% We hypothesized that variational Bayes approach can avoid over-fitting on a small amount of training samples because it has been demonstrated that the generalization ability of the predictive models with stochasticity is better than without stochasticity large amount of samples \cite{Babaeizadeh:2017,Denton:2018} and a stochastic sampling process will provide .
We compared the deterministic and stochastic predictive models to verify the effect of variational Bayes in the predictive coding framework. 
The stochastic model sampled initial states from reparameterized normal distributions in Eq. (\ref{eq:reparameterization}).
In the deterministic model, initial states were parameterized only using a mean without a standard deviation, because there is no sampling process for initial states.
The losses for training and planning of the deterministic model were the same as the stochastic model, but the Kullback-Leibler divergence between prior and posterior initial states was computed with a fixed standard deviation set to 1. During the planning phase, error regression was repeated over 500 epochs without sampling.

In qualitative analysis, we discovered that the deterministic model had difficulty finding plausible visuomotor sequences, which indicates the lack of generalization.
Fig. \ref{fig:qualitative_comparision}(c) shows that the robot disappeared while picking the first block and the robot's picking movements were not accurate. 
Also, the motor prediction little resembled the target motor sequence. 
In quantitative analysis, the stochastic model significantly outperformed the deterministic model as shown in Fig. \ref{fig:quantitative_comparision}.
Note that, in the case of the (stochastic) model without background memory, the model failed to plan the correct visuomotor sequence, but the predicted visuomotor sequence seemed at least plausible.
These results indicate that the variational predictive coding framework provides better generalization capability than the deterministic one.

% stochastic imporves generalization?

\section{Conclusion}
In this paper, we proposed a neural network model under variational predictive coding incorporating top-down visual attention and visual working memory for goal-directed, long-term planning through mental simulation.  
The experimental results for the task of block stacking demonstrated that the proposed model was able to plan long-term, high-dimensional visuomotor sequences by Bayesian inference in low-dimensional latent intentional space. 
% Furthermore, we compared between the proposed model and ablated variants that excluded visual working memory or variational Bayes in a qualitative and quantitative manner.
Furthermore, the proposed model clearly outperformed ablated variants that excluded visual working memory or variational Bayes in a qualitative and quantitative manner.
Our results showed that the proposed model self-organized smooth pursuit-like movements that tracked the end effector, mentally manipulated visuospatial information stored in visual working memory, and achieved remarkable generalization ability in a sample-efficient manner by employing variational Bayes.
% However, since the current study focused on planning in an offline manner, the proposed model was hard to cope with failures that occurred during action execution. 
The future study should involve with replanning during action execution in an online manner because the current study focused on planning in an offline manner.

\addtolength{\textheight}{-12cm}   % This command serves to balance the column lengths

%%%%%%%%%%%%%%%%%%%%%%%%%%%%%%%%%%%%%%%%%%%%%%%%%%%%%%%%%%%%%%%%%%%%%%%%%%%%%%%%

%%%%%%%%%%%%%%%%%%%%%%%%%%%%%%%%%%%%%%%%%%%%%%%%%%%%%%%%%%%%%%%%%%%%%%%%%%%%%%%%



% \section*{ACKNOWLEDGMENT}

% The preferred spelling of the word �acknowledgment� in America is without an �e� after the �g�. Avoid the stilted expression, �One of us (R. B. G.) thanks . . .�  Instead, try �R. B. G. thanks�. Put sponsor acknowledgments in the unnumbered footnote on the first page.

%%%%%%%%%%%%%%%%%%%%%%%%%%%%%%%%%%%%%%%%%%%%%%%%%%%%%%%%%%%%%%%%%%%%%%%%%%%%%%%%

% References are important to the reader; therefore, each citation must be complete and correct. If at all possible, references should be commonly available publications.
\bibliographystyle{IEEEtran}
\bibliography{references}

\begin{thebibliography}{10}
\providecommand{\url}[1]{#1}
\csname url@samestyle\endcsname
\providecommand{\newblock}{\relax}
\providecommand{\bibinfo}[2]{#2}
\providecommand{\BIBentrySTDinterwordspacing}{\spaceskip=0pt\relax}
\providecommand{\BIBentryALTinterwordstretchfactor}{4}
\providecommand{\BIBentryALTinterwordspacing}{\spaceskip=\fontdimen2\font plus
\BIBentryALTinterwordstretchfactor\fontdimen3\font minus
  \fontdimen4\font\relax}
\providecommand{\BIBforeignlanguage}[2]{{%
\expandafter\ifx\csname l@#1\endcsname\relax
\typeout{** WARNING: IEEEtran.bst: No hyphenation pattern has been}%
\typeout{** loaded for the language `#1'. Using the pattern for}%
\typeout{** the default language instead.}%
\else
\language=\csname l@#1\endcsname
\fi
#2}}
\providecommand{\BIBdecl}{\relax}
\BIBdecl

\bibitem{Finn:2017}
C.~Finn and S.~Levine, ``Deep visual foresight for planning robot motion,'' in
  \emph{IEEE International Conference on Robotics and Automation (ICRA)}, 2017.

\bibitem{Choi:2018}
M.~Choi, T.~Matsumoto, M.~Jung, and J.~Tani, ``Generating goal-directed
  visuomotor plans based on learning using a predictive coding type deep
  visuomotor recurrent neural network model,'' \emph{CoRR}, vol.
  abs/1803.02578, 2018.

\bibitem{Nair:2018}
A.~V. Nair, V.~Pong, M.~Dalal, S.~Bahl, S.~Lin, and S.~Levine, ``Visual
  reinforcement learning with imagined goals,'' in \emph{Advances in Neural
  Information Processing Systems (NIPS)}, 2018.

\bibitem{Levine:2018}
S.~Levine, P.~Pastor, A.~Krizhevsky, J.~Ibarz, and D.~Quillen, ``Learning
  hand-eye coordination for robotic grasping with deep learning and large-scale
  data collection,'' \emph{The International Journal of Robotics Research},
  vol.~37, no. 4-5, pp. 421--436, 2018.

\bibitem{Grezes:2001}
J.~Grèzes and J.~Decety, ``Functional anatomy of execution, mental simulation,
  observation, and verb generation of actions: A meta-analysis,'' \emph{Human
  Brain Mapping}, vol.~12, no.~1, pp. 1--19, 2001.

\bibitem{Pezzulo:2014a}
G.~Pezzulo, M.~A. van~der Meer, C.~S. Lansink, and C.~M. Pennartz, ``Internally
  generated sequences in learning and executing goal-directed behavior,''
  \emph{Trends in Cognitive Sciences}, vol.~18, no.~12, pp. 647--657, 2014.

\bibitem{Rao:1999}
R.~P.~N. Rao and D.~H. Ballard, ``Predictive coding in the visual cortex: a
  functional interpretation of some extra-classical receptive-field effects,''
  \emph{Nature Neuroscience}, vol.~2, pp. 79--87, 1999.

\bibitem{Friston:2010}
K.~Friston, ``The free-energy principle: a unified brain theory?'' \emph{Nature
  Reviews Neuroscience}, vol.~11, pp. 127--138, 2010.

\bibitem{Coltheart:1980}
M.~Coltheart, ``Iconic memory and visible persistence,'' \emph{Perception {\&}
  Psychophysics}, vol.~27, no.~3, pp. 183--228, 1980.

\bibitem{Baddeley:2003}
A.~Baddeley, ``Working memory: looking back and looking forward,'' \emph{Nature
  Reviews Neuroscience}, vol.~4, pp. 829--839, 2003.

\bibitem{Sims:1997}
V.~K. Sims and M.~Hegarty, ``Mental animation in the visuospatial sketchpad:
  Evidence from dual-task studies,'' \emph{Memory {\&} Cognition}, vol.~25,
  no.~3, pp. 321--332, 1997.

\bibitem{Roelfsema:2016}
P.~Roelfsema and F.~{De Lange}, ``Early visual cortex as a multiscale cognitive
  blackboard.'' \emph{Annual Review of Vision Science}, vol.~2, pp. 131--151,
  2016.

\bibitem{Kingma:2013}
D.~P. Kingma and M.~Welling, ``Auto-encoding variational bayes,'' \emph{CoRR},
  vol. abs/1312.6114, 2013.

\bibitem{Norman:2002}
J.~Norman, ``Two visual systems and two theories of perception: An attempt to
  reconcile the constructivist and ecological approaches,'' \emph{Behavioral
  and Brain Sciences}, vol.~25, no.~1, pp. 73--96, 2002.

\bibitem{Sheth:2016}
B.~R. Sheth and R.~Young, ``Two visual pathways in primates based on sampling
  of space: Exploitation and exploration of visual information,''
  \emph{Frontiers in integrative neuroscience}, vol.~10, p.~37, 2016.

\bibitem{Shi:2015}
X.~Shi, Z.~Chen, H.~Wang, D.-Y. Yeung, W.-k. Wong, and W.-c. Woo,
  ``Convolutional lstm network: A machine learning approach for precipitation
  nowcasting,'' in \emph{Advances in Neural Information Processing Systems
  (NIPS)}, 2015.

\bibitem{Hochreiter:1997}
S.~Hochreiter and J.~Schmidhuber, ``Long short-term memory,'' \emph{Neural
  Computation}, vol.~9, no.~8, pp. 1735--1780, 1997.

\bibitem{Jaderberg:2015}
M.~Jaderberg, K.~Simonyan, A.~Zisserman, and k.~kavukcuoglu, ``Spatial
  transformer networks,'' in \emph{Advances in Neural Information Processing
  Systems (NIPS)}, 2015.

\bibitem{Eslami:2016}
S.~M.~A. Eslami, N.~Heess, T.~Weber, Y.~Tassa, D.~Szepesvari, k.~kavukcuoglu,
  and G.~E. Hinton, ``Attend, infer, repeat: Fast scene understanding with
  generative models,'' in \emph{Advances in Neural Information Processing
  Systems (NIPS)}, 2016.

\bibitem{Nishimoto:2004}
R.~Nishimoto and J.~Tani, ``Learning to generate combinatorial action sequences
  utilizing the initial sensitivity of deterministic dynamical systems,''
  \emph{Neural Networks}, vol.~17, no.~7, pp. 925--933, 2004.

\bibitem{Shima:2006}
K.~Shima, M.~Isoda, H.~Mushiake, and J.~Tanji, ``Categorization of behavioural
  sequences in the prefrontal cortex,'' \emph{Nature}, vol. 445, pp. 315--318,
  2006.

\bibitem{Yamashita:2008}
Y.~Yamashita and J.~Tani, ``Emergence of functional hierarchy in a multiple
  timescale neural network model: A humanoid robot experiment,'' \emph{PLOS
  Computational Biology}, vol.~4, no.~11, pp. 1--18, 2008.

\bibitem{Babaeizadeh:2017}
M.~Babaeizadeh, C.~Finn, D.~Erhan, R.~H. Campbell, and S.~Levine, ``Stochastic
  variational video prediction,'' \emph{CoRR}, vol. abs/1710.11252, 2017.

\bibitem{Denton:2018}
E.~Denton and R.~Fergus, ``Stochastic video generation with a learned prior,''
  \emph{CoRR}, vol. abs/1802.07687, 2018.

\bibitem{Ba:2016}
L.~J. Ba, R.~Kiros, and G.~E. Hinton, ``Layer normalization,'' \emph{CoRR},
  vol. abs/1607.06450, 2016.

\bibitem{Kingma:2014}
D.~P. Kingma and J.~Ba, ``Adam: {A} method for stochastic optimization,''
  \emph{CoRR}, vol. abs/1412.6980, 2014.

\bibitem{Pascanu:2013}
R.~Pascanu, T.~Mikolov, and Y.~Bengio, ``On the difficulty of training
  recurrent neural networks,'' in \emph{International Conference on Machine
  Learning (ICML)}, 2013.

\bibitem{Wijesinghe:2018}
L.~P. Wijesinghe, J.~Triesch, and B.~E. Shi, ``Robot end effector tracking
  using predictive multisensory integration,'' \emph{Frontiers in
  Neurorobotics}, vol.~12, p.~66, 2018.

\bibitem{AXLee:2018}
A.~X. Lee, R.~Zhang, F.~Ebert, P.~Abbeel, C.~Finn, and S.~Levine, ``Stochastic
  adversarial video prediction,'' \emph{CoRR}, vol. abs/1804.01523, 2018.

\bibitem{Zhang:2018}
R.~Zhang, P.~Isola, A.~A. Efros, E.~Shechtman, and O.~Wang, ``The unreasonable
  effectiveness of deep features as a perceptual metric,'' in \emph{IEEE
  Conference on Computer Vision and Pattern Recognition (CVPR)}, 2018.

\bibitem{Sakoe:1978}
H.~Sakoe and S.~Chiba, ``Dynamic programming algorithm optimization for spoken
  word recognition,'' \emph{IEEE Transactions on Acoustics, Speech, and Signal
  Processing}, vol.~26, no.~1, pp. 43--49, 1978.

\bibitem{Ghahramani:2015}
Z.~Ghahramani, ``Probabilistic machine learning and artificial intelligence,''
  \emph{Nature}, vol. 521, pp. 452--459, 2015.

\end{thebibliography}

\end{document}